\documentclass{article}

    \usepackage[preprint]{neurips_data_2024}

\usepackage[utf8]{inputenc} 
\usepackage[T1]{fontenc}    
\usepackage{hyperref}       
\usepackage{url}            
\usepackage{booktabs}       
\usepackage{amsfonts}       
\usepackage{nicefrac}       
\usepackage{microtype}      
\usepackage{xcolor}         
\usepackage{amsmath}
\usepackage{algorithm}
\usepackage{tabularx}
\usepackage{algpseudocode}
\usepackage{graphicx}
\usepackage{csquotes}

\title{BanglishRev: A Large-Scale Bangla-English and Code-mixed Dataset of Product Reviews in E-Commerce}

\author{%
  Mohammad Nazmush Shamael \\
  United International University\\
  United City, Madani Ave, Dhaka, Bangladesh \\
  \texttt{mshamael2320013@mscse.uiu.ac.bd} \\
  \And
  Sabila Nawshin \\
  Indiana University Bloomington\\
  Bloomington, Indiana, USA\\
  \texttt{snawshin@iu.edu} \\
  \And
  Swakkhar Shatabda \\
  BRAC University\\
  Merul Badda, Dhaka, Bangladesh \\
  \texttt{swakkhar.shatabda@bracu.ac.bd} \\
  \And
  Salekul Islam \\
  North South University\\
  Bashundhara, Dhaka, Bangladesh \\
  \texttt{salekul.islam@northsouth.edu} \\
}

\begin{document}

\maketitle

\begin{abstract}
This work presents the BanglishRev Dataset, the largest e-commerce product review dataset to date for reviews written in Bengali, English, a mixture of both and Banglish, Bengali words written with English alphabets. The dataset comprises of 1.74 million written reviews from 3.2 million ratings information collected from a total of 128k products being sold in online e-commerce platforms targeting the Bengali population. It includes an extensive array of related metadata for each of the reviews including the rating given by the reviewer, date the review was posted and date of purchase, number of likes, dislikes, response from the seller, images associated with the review etc. With sentiment analysis being the most prominent usage of review datasets, experimentation with a binary sentiment analysis model with the review rating serving as an indicator of positive or negative sentiment was conducted to evaluate the effectiveness of the large amount of data presented in BanglishRev for sentiment analysis tasks. A BanglishBERT model is trained on the data from BanglishRev with reviews being considered labeled positive if the rating is greater than 3 and negative if the rating is less than or equal to 3. The model is evaluated by being testing against a previously published manually annotated dataset for e-commerce reviews written in a mixture of Bangla, English and Banglish. The experimental model achieved an exceptional accuracy of 94\% and F1 score of 0.94, demonstrating the dataset's efficacy for sentiment analysis. Some of the intriguing patterns and observations seen within the dataset and future research directions where the dataset can be utilized is also discussed and explored. The dataset can be accessed through \href{https://huggingface.co/datasets/BanglishRev/bangla-english-and-code-mixed-ecommerce-review-dataset}{https://huggingface.co/datasets/BanglishRev/bangla-english-and-code-mixed-ecommerce-review-dataset}.
\end{abstract}

\section{Introduction} \label{Introduction}

\begin{figure}[t]
\centering
\includegraphics[width=0.90\textwidth]{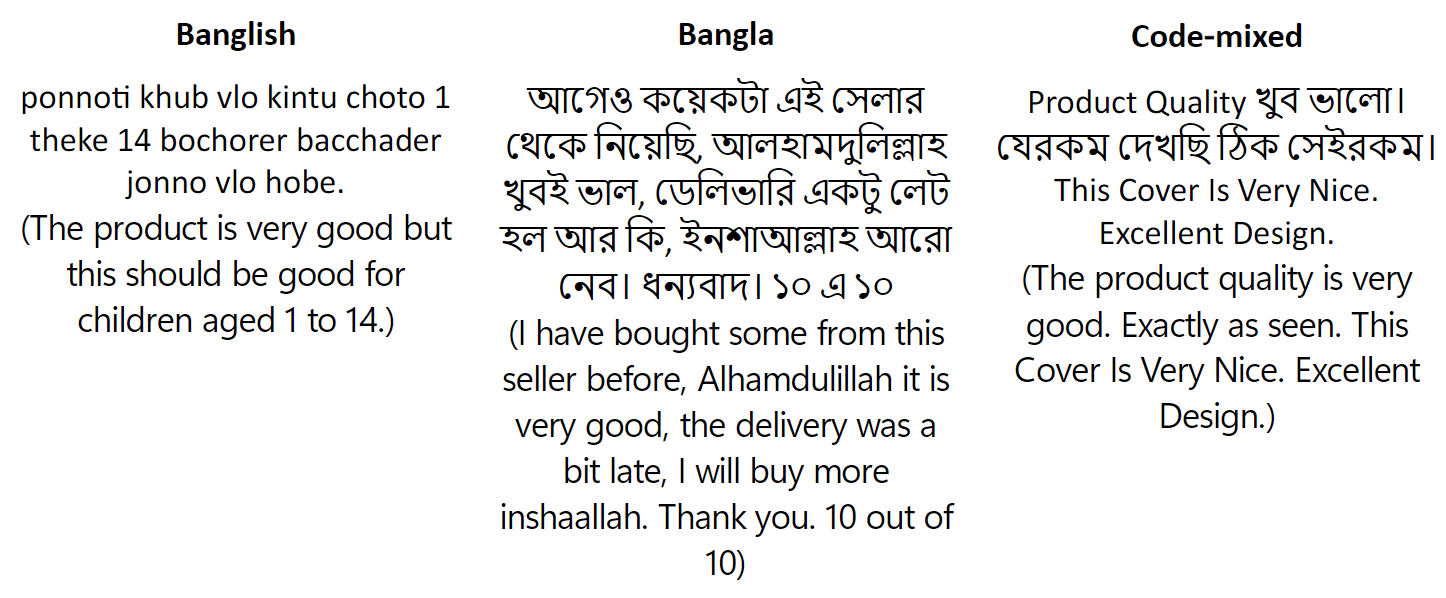}
\caption{Sample Banglish, Bangla and Code-mixed texts and their corresponding English translation} 
\label{fig:word_sample}
\end{figure}

Bangla, an Indo-Aryan language originating from the Indo-European languague family is the sixth most spoken language in the world \citep{klaiman2018bengali} with more than 200 million people in the Bengal region of South Asia using it as their mother tongue. The e-commerce sector targeting this substantial population, specifically Bangladesh, has witnessed significant expansion in the past decade. Platforms like Rokomari.com, Daraz, Othoba.com, Pickaboo, Chaldal etc. has been spearheading online retail in their respective domains \citep{mohiuddin2014overview}. The primary form of expression for this customer base is Bengali, English, a mixture of Bengali and English words in the same sentence/utterance (known as code mixing or mixed code language) and Bengali words written in English alphabets, commonly known as Banglish.  Figure \ref{fig:word_sample} shows a sample Bengali, Code-mixed (mixture of Bengali and English words in the same statement) and Banglish sentence with their translations to aid understanding. 

It is imperative to better understand the evolving consumer preferences and customer behaviors in the changing e-commerce landscape to further expand this dynamic market. Sentiment analysis of the Bangla language, specifically, has seen increasing use of e-commerce review data in recent years \citep{akter2021bengali, hossain2022sentiment, shafin2020product, munna2020sentiment, purba2022hybrid, sharmin2021attention, karmakar2022sentiment}. Quite a few review datasets \citep{rashid2024comprehensive, hossain2021rating,shanto2023mining, akter2021bengali, hossain2022sentiment} have been curated from one of the most prominent e-commerce platforms in Bangladesh, Daraz\footnote{\href{https://www.daraz.com.bd/}{https://www.daraz.com.bd/}}. The largest one available contains 78k reviews \citep{rashid2024comprehensive} with to little to no metadata associated with the reviews available in the dataset and is specifically curated for sentiment analysis. While sentiment analysis of product reviews can help analyze the user experience of different products, the metadata related with those reviews is essential in getting a bigger picture of the whole market.

This work introduces the BanglishRev Dataset, the largest e-commerce product dataset to date for the Bengali customer base. It contains 1.74 million reviews from a total of 3.2 million rating information, written in a mixture of Bangla, English and Banglish. The data was collected from a wide array 128k products available in the e-commerce platform Daraz Bangladesh. For each of the reviews in the dataset, a broad spectrum of associated metadata such as the rating of the product provided by the reviewer, the date of the review and purchase, the number of likes and dislikes, responses from sellers and images related to the review is also included in the dataset. BanglishRev is the first dataset curated in the context of the Bengali population that includes the images associated with the reviews. There are meta-data available for each of the product as well. Since sentiment analysis is one of the major use-cases for Bengali review datasets, rating information present in the dataset is used as labels for an supervised model to train a BanglishBert \citep{bhattacharjee2021banglabert} model on BanglishRev and it is tested against the manually labeled 78k dataset curated by \cite{rashid2024comprehensive}. Multiple experiments with different thresholds of ratings serving as indicators of positive and negative sentiment have been conducted in this work. The experimental model exhibited promising results, underscoring the dataset's suitability for training machine learning models in sentiment analysis tasks. Future directions to be taken and observations made from the dataset is also discussed in the later sections.


\section{Related works} \label{Related works}
Several datasets specifically curated for Bengali population has emerged in the past few years, this section discusses the datasets and related works. Table \ref{table:review_comparison} provides an brief overview of the existing Bengali review datasets. 
\begin{table} [tbp]
\centering
\small
\caption{Bangla review datasets in literature}
\begin{tabularx}{\textwidth}{l l X}
\toprule
        Author & Dataset size & Features \\  
        \midrule
        
        \cite{rashid2024comprehensive} & 78130 & Product Rating, Review Text, Product Name, Product Category, Emotion, Sentiment, Data Source\\ 
        
        \cite{shanto2023mining} & 1000 & Review, Review Type, Product Type\\ 
        
        \cite{hossain2021rating} & 7874 & Reviewer's username, Product Category and sub-category, Product Type, Product Name, Review, Review Date, Rating\\ 
        
        \cite{akter2021bengali} & 7905 & Reviewer's username, Product Category and sub-category, Product Type, Product Name, Review, Rating, Sentiment\\ 
        
        \cite{hossain2022sentiment} & 7181 & Review Text, Language Used, Sentiment\\ 
\bottomrule
\end{tabularx}
\label{table:review_comparison}
\end{table}

The most recent and prominent dataset for Bangla reviews was curated by \cite{rashid2024comprehensive}. The dataset consisted 78,130 reviews from Daraz and Pickaboo across 152 different product categories, with 28,912 Bangla reviews and 49,218 English reviews. The reviews were manually annotated. While much larger than the previously available datasets, this dataset is specifically curated for sentiment analysis and lacks any other data associated with the reviews or the products.

\cite{shanto2023mining} collected 1000 Bangla reviews from Daraz across 5 product categories. The reviews in this dataset are also manually annotated in both Binary (Positive / Negative) and Multi-Class(Very Positive, Positive, Negative, and Very Negative) sentiment. The dataset also contains English translations of the Bangla reviews. 

7,874 Bangla reviews from Daraz were gathered by \cite{hossain2021rating}. Their dataset included information about the category, sub-category, username of the reviewer, product type, product name and review date and remains unlabeled for sentiment analysis tasks. The authors curated the dataset to predict the ratings of products. 

\cite{akter2021bengali} amassed 7,905 reviews in Bangla. The dataset included information about the reviewer's username, the category and sub category of the product, product type and product name, review date, rating given by the reviewer, and manually labeled sentiments. The sentiments are deduced from the rating, in a manner similar to ours. Ratings higher than 3 are deemed positive, a rating of 3 is considered neutral, and ratings less than 3 are interpreted as negative sentiment in the dataset. While counting similarly extensive array of metadata, the dataset is considerably smaller than BanglishRev. 

\cite{hossain2022sentiment} curated a dataset consisting of 7,181 reviews. 3,964 reviews among those are in Bangla, 2,059 reviews are in English, and 1,161 reviews are in romanized Bangla. Each review in the dataset is manually annotated by the authors with one of five sentiments: Very Positive, Positive, Neutral, Negative, or Very Negative.

Aside from Bangla, \cite{sutoyo2022prdect} curated an Indonesian product reviews dataset with 5400 reviews, \cite{ganganwar2023enhanced} curated a review dataset for Hindi, \cite{noor2019sentiment} worked on a Urdu review dataset with 20.286K reviews and \cite{keung2020multilingual} presented an multilingual Amazon review dataset among many others.

\section{Data Collection and Analysis} \label{Data Collection and Analysis}
\subsection{Data Collection}
The data collection methodology for curating BanglishRev is discussed in this section. The methodology for constructing the review dataset can be divided into three main parts, category link collection, product url collection and review details collection. Selenium\footnote{\href{https://www.selenium.dev/}{https://www.selenium.dev/}} and beautifulsoup\footnote{\href{https://beautiful-soup-4.readthedocs.io/en/latest/}{https://beautiful-soup-4.readthedocs.io/en/latest/}} python libraries have been used for the automated product information scraping in this work. Each of the steps of the in the data collection procedure is discussed in details below. 



\textbf{Category link collection}: Daraz has a category selection menu that follows a 3-tier tree structure. Each individual product is labeled under a category that serves as one of the eight root categories from the 3-tier tree structure, then a sub-category with the root category as a parent and a sub-sub-category with the sub-category as the parent. The categories in the topmost level of the structure is henceforth addressed as the 'Root Category', the second tier the 'Parent Category' and the last layer simply as 'Category'. In order to scrape the product categories and their corresponding links, the homepage of Daraz is visited in the first step and the Root Categories and their corresponding information is extracted manually. This information is then used to extract the Parent Category information and Category information automatically. Finally, the extracted files are stored in a CSV file. 

\textbf{Product URL collection}: In each category pages of Daraz, there are product cards that include various information of the product. The previously stored category link are iterated for collecting the product URLs. The last page number from the page is first scarped and used to generate the page links. The product URLs of the products are then extracted with ratings from all pages of the category. Afterwards, the algorithm moves on to the next category and start all over again. This process is continued until all category pages are scraped. Finally, the product information is saved in a JSON file.

\textbf{Review information collection}: In the review section of each product, there are multiple pages of reviews with each review containing various information like user rating, review, likes, dislikes, date of review and purchase, review images etc. For collecting the product reviews, the product pages stored in the JSON file of product URLs are analyzed. Each product page has multiple review pages containing reviews. The algorithm iterates through all of the review pages and extracts the review information. This process is continued until all product pages are scraped. Finally the review information is saved in another JSON file.

The data in the JSON file from the review information collection is anonymized in the final version of BanglishRev and information like product ID and user ID are replaced with unique numbers. The resulting dataset is a detailed review dataset containing reviews for 128,543 products out of more than 1 million products, most among which did not have any rating or review information associated with it.

\subsection{Dataset Analysis}
A brief analysis of the data present in the BanglishRev dataset is provided in this section.

\begin{figure}[htbp]
    \centering
    \begin{minipage}[b]{0.47\textwidth}
        \centering
        \includegraphics[width=\textwidth]{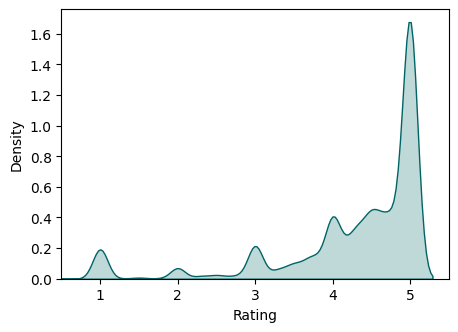}
        \caption{Density plot of product rating}
        \label{fig:rating dist}
    \end{minipage}
    \hfill
    \begin{minipage}[b]{0.47\textwidth}
        \centering
        \includegraphics[width=\textwidth]{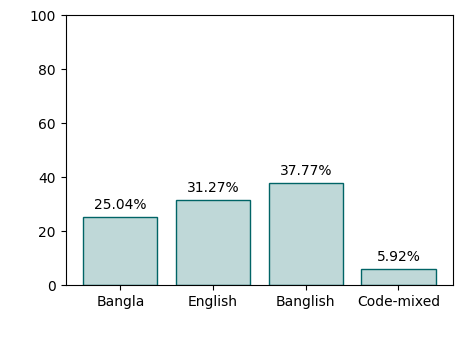}
        \caption{Distribution of reviews by Language}
        \label{fig:review type dist}
    \end{minipage}

\end{figure}

A total of 3,239,811 rating information is collected in BanglishRev, out of which 1,747,043 contain written reviews from the user. Each rating information consists of the buyer ID, rating of the product, review content, review date, buying date, likes, dislikes, seller reply and information about the images (links) associated with the review. Additionally, each product has its corresponding average rating, score distribution, number of reviews and category information. The distribution of the average product rating is shown in Figure \ref{fig:rating dist}. As observed in Figure \ref{fig:rating dist}, a major portion of the products with reviews have 5 star ratings.

In order to get a better grasp of the distribution of review types, the reviews are categorized into four different categories by language, namely Bangla reviews, English reviews, code-mixed reviews and Banglish reviews. The distribution of the reviews by the Language category can be seen in Figure \ref{fig:review type dist}. As can be seen, 37.7\% of the total reviews are written in Banglish, 21.2\% in English, 25.04\% in English and only 5.9\% in code-mixed language. 

\begin{figure}[tp]
    \centering
    \begin{minipage}[b]{0.535\textwidth}
        \centering
        \includegraphics[width=\textwidth]{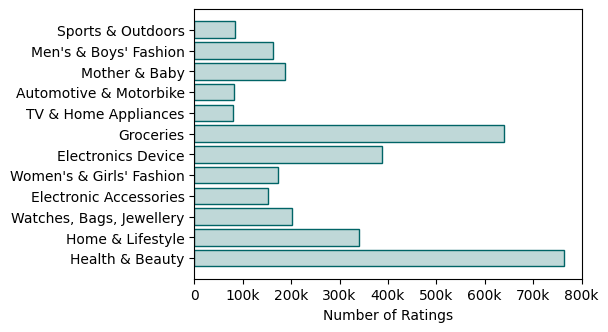}
        \caption{Rating distribution per root category}
        \label{fig:Number of Ratings per category}
    \end{minipage}
    \hfill
    \begin{minipage}[b]{0.415\textwidth}
        \centering
        \includegraphics[width=\textwidth]{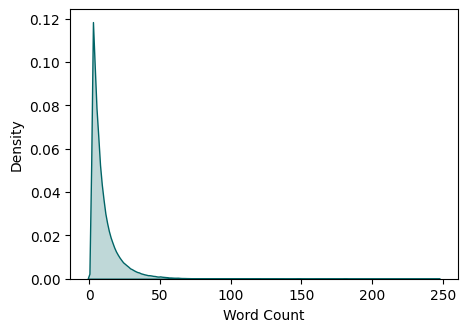}
        \caption{Word count distribution}
        \label{fig:Word count density}
    \end{minipage}
\end{figure}

The rating distribution over the 12 root categories is shown in Figure \ref{fig:Number of Ratings per category}. Here, the root category with the most number of reviews is the Heath and Beauty Category with Groceries trailing close to it. Figure \ref{fig:Word count density} shows a density plot of the review lengths observed in the dataset. A considerable portion of the dataset contains only 2 word reviews (e.g. "Good product", "Very good", "Nice product"), which may be the case because review rating is negatively correlated to review length \citep{ghasemaghaei2018reviews} and BanglishRev contains an overwhelming number of 5 star ratings.

\begin{figure}[tp]
    \centering
    \begin{minipage}[b]{0.45\textwidth}
        \centering
        \includegraphics[width=\textwidth]{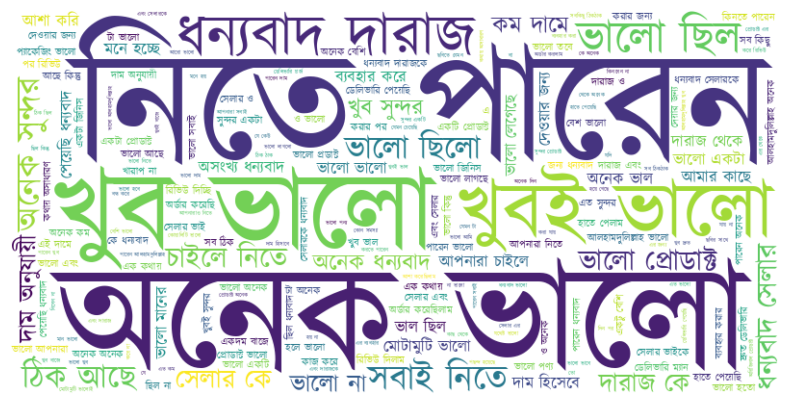}
        \caption{Most used words in Bangla}
        \label{fig:Bangla wordcloud.png}
    \end{minipage}
    \hfill
    \begin{minipage}[b]{0.45\textwidth}
        \centering
        \includegraphics[width=\textwidth]{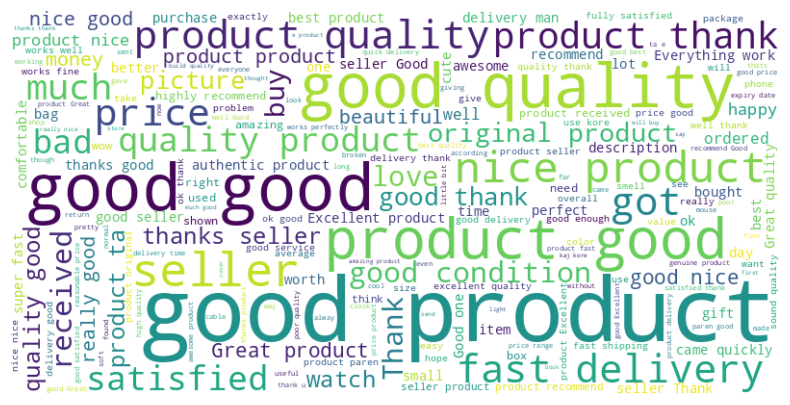}
        \caption{Most used words in English}
        \label{fig:English wordcloud.png}
    \end{minipage}
\end{figure}

Regular expressions are used to identify the Bangla alphabets and to create a word cloud that identifies the most used words in Bangla. \ref{fig:Bangla wordcloud.png} shows the resulting word cloud. The most used words translates to "Very good" and "Take this product".  The word that came up the most in the English reviews is "good", with various other words appearing alongside ("product", "quality", "price" etc.). Figure \ref{fig:English wordcloud.png} visualizes those words with the most frequency. However, the approach used for Bangla could not be used in identifying English words as both English and Banglish reviews are written using the English alphabets. For this reason, the NLTK\footnote{\href{https://www.nltk.org/}{https://www.nltk.org/}} library is used to identify English words in the reviews. All reviews either written fully using Bangla or using English are identified at first in this method, and then the code-mixed reviews where both Bangla and English words are used in the same review are identified. The remaining reviews are a combination of Banglish review and code-mixed Banglish reviews where a combination of Banglish and English words are used in the review. In total, we found 437470 or 25.04\% of the reviews to be completely in Bangla, 546363 or 31.27\% reviews to be in complete English, 103436 or 5.92\% to be in Bangla-English code-mixed and 659774 or 37.77\% reviews to be in Banglish. It should be noted that this approach classifies English reviews that have misspelled words in it as Banglish reviews. Further fine tuning of the conditions is required to classify such reviews as there are no straight forward way to separate them.

\section{Experiments and Discussion} \label{Experiments and Discussion}
\subsection{Effectiveness of rating as sentiment classification label}
In the review of the existing literature related to e-commerce reviews in Bangladesh, it was apparent that a large portion of the datasets and existing work focused on sentiment analysis of the product reviews. As the proposed BanglishRev dataset is very large and it is challenging to manually annotate it with sentiment labels, rating can be a suitable alternative for simple sentiment classification \citep{nguyen2014sentiment, singh2013sentiment}. In this section, experimentation on the effectiveness of BanglishRev in training a binary sentiment (positive/negative) classification model is provided.

\subsubsection{Implementation Details}\label{Implementation Details}

\textbf{Experimental setup}: Google Colaboratory\footnote{\href{https://colab.research.google.com/}{https://colab.research.google.com/}} has been used as main platformfor this experiment. A runtime with NVIDIA A100 GPU with 40GB of GPU RAM, Intel(R) Xeon(R) CPU @ 2.20GHz CPU, and 83.5GB RAM was used in training and testing the models, as training the model with the BanglishRev dataset proved to be very reliant on larger GPU RAM capacity.\\

\textbf{Implemetation:} When dealing with multilingual and code-mixed text like the user reviews of the dataset, a robust normalization process for the data is necessary for reducing noise and improve tokenization. In order to handle the various types of review in the dataset, a pipeline was created for normalizing the text. First, emojis and non-terminal punctuation are removed from the review and then it is determined whether the review is in Bangla, English, Banglish or code-mixed language. This is done by counting the number of English and Bangla words in the review. The NLTK library is used to check if a word written in the English alphabet is a valid English word and words written in the Bangla alphabet are identified as Bangla. The words that are neither identified as Bangla or English are considered as Banglish and are converted into its phonetic equivalent of Bangla with the avro\footnote{\href{https://pypi.org/project/avro-py/}{https://pypi.org/project/avro-py/}} library. Finally, any extra white-space is removed. Details about the normalization process has been included in the supplementary material. A BanglishBERT \citep{bhattacharjee2021banglabert}, BERT-based \citep{devlin2018bert} model designed to bridge the linguistic gap between Bangla and English, is used in this work for the sentiment classification task. It utilizes the ELECTRA \citep{clark2020electra} discriminator framework and is trained with the Replaced Token Detection (RTD) objective. BanglishBERT demonstrates remarkable zero-shot cross-lingual transfer capabilities across various natural language processing (NLP) tasks in the Bangla domain. The max token size is set to 256 for tokenization in this implementation. Adam optimizer is utilized to train the models for 3 epochs with a batch size of 128 and learning rate of 0.00005. 

\subsubsection{Experimental Results}
Three experiments were conducted to measure the effectiveness of rating as sentiment classification label. The experiments are as follows:

\noindent\textbf{Experiment-1:} A baseline model trained and tested on the dataset curated by \cite{rashid2024comprehensive}

\noindent\textbf{Experiment-2:} Model trained on BanglishRev with all reviews with a rating of 4 or more considered to be positive and 3 or less considered to be negative.

\noindent\textbf{Experiment-3:} Model trained on BanglishRev with all reviews with a rating of 3 or more considered to be positive and 2 or less considered to be negative.

The dataset curated by \citep{rashid2024comprehensive} is the largest e-commerce review dataset currently available in literature. The dataset was manually annotated for positive and negative sentiment, and is a fitting dataset for setting a baseline for the experiments. In this baseline experiment, experiment 1, 80\% of the data in the dataset by \cite{rashid2024comprehensive} is used for training (based on positive/negative sentiment labels) and 20\% data for testing. An accuracy of 93\% is obtained on the test data, but the model performed considerably worse (precision of 0.69, recall of 0.89 and F1-score of 0.89 when classifying negative sentiments). Detailed result can be found on Table \ref{table:ex1e3}.

Experiment 1 and 2 are conducted with the BanglishRev dataset being used for training the BanglishBERT model and the dataset curated by \cite{rashid2024comprehensive} being used for testing. Both experiments are ran for 1 epoch at first and tested using the dataset curated by \cite{rashid2024comprehensive}. After 1 epoch in experiment 2, an accuracy of 95\% is observed with a precision and recall of 0.85 and 0.73 respectively when classifying negative sentiments whereas 93\% accuracy is observed in experiment 3 with a horrible recall score of 0.55 when classifying negative sentiments. The detailed result of both experiment 2 and 3 after 1 epoch can be found on Table \ref{table:ex2e1} and Table \ref{table:ex3e1}.

\begin{table}[t]
\small
\centering
\begin{minipage}{0.46\textwidth}
    \centering
    \caption{Experiment 1 results after 3 Epoch}
    \label{table:ex1e3}
    \begin{tabular}{lccc}
    \toprule
    \textbf{Class} & \textbf{Precision} & \textbf{Recall} & \textbf{F1-score} \\
    \midrule
    Positive & 0.98 & 0.94 & 0.96 \\
    Negative & 0.69 & 0.89 & 0.78 \\
    \textbf{Accuracy} & & & 0.93\\
    \midrule
    \textbf{Macro avg} & 0.84 & 0.91 & 0.87 \\
    \textbf{Weighted avg} & 0.94 & 0.93 & 0.93 \\
    \bottomrule
    \end{tabular}
\end{minipage}
\hspace{0.05\textwidth}
\begin{minipage}{0.46\textwidth}
    \centering
    \caption{Experiment 2 results after 3 Epoch}
    \label{table:ex2e3}
    \begin{tabular}{lccc}
    \toprule
    \textbf{Class} & \textbf{Precision} & \textbf{Recall} & \textbf{F1-score} \\
    \midrule
    Positive & 0.96 & 0.98 & 0.97 \\
    Negative & 0.84 & 0.72 & 0.78 \\
   \textbf{Accuracy} & & & 0.94\\
    \midrule
    \textbf{Macro avg} & 0.90 & 0.85 & 0.87 \\
    \textbf{Weighted avg} & 0.94 & 0.94 & 0.94 \\
    \bottomrule
    \end{tabular}
\end{minipage}
\end{table}

\begin{table}[t]
\small
\centering
\begin{minipage}{0.46\textwidth}
    \centering
    \caption{Experiment 3 results after 1 Epoch}
    \label{table:ex3e1}
    \begin{tabular}{lccc}
    \toprule
    \textbf{Class} & \textbf{Precision} & \textbf{Recall} & \textbf{F1-score} \\
    \midrule
    Positive & 0.93 & 0.99 & 0.96 \\
    Negative & 0.90 & 0.55 & 0.69 \\
    \textbf{Accuracy} & & & 0.93\\
    \midrule
    \textbf{Macro avg} & 0.92 & 0.77 & 0.82 \\
    \textbf{Weighted avg} & 0.93 & 0.93 & 0.92 \\
    \bottomrule
    \end{tabular}
\end{minipage}
\hspace{0.05\textwidth}
\begin{minipage}{0.46\textwidth}
    \centering
    \caption{Experiment 2 results after 1 Epoch}
    \label{table:ex2e1}
    \begin{tabular}{lccc}
    \toprule
    \textbf{Class} & \textbf{Precision} & \textbf{Recall} & \textbf{F1-score} \\
    \midrule
    Positive & 0.96 & 0.98 & 0.97 \\
    Negative & 0.85 & 0.73 & 0.79 \\
    \textbf{Accuracy} & & & 0.95\\
    \midrule
    \textbf{Macro avg} & 0.91 & 0.86 & 0.88 \\
    \textbf{Weighted avg} & 0.94 & 0.95 & 0.94 \\
    \bottomrule
    \end{tabular}
\end{minipage}
\end{table}

Due to the training process using the BanglishRev dataset being computationally very expensive and limitations in computational resources bought in Google Colab and the poor results gained, experiment 3 was not conducted further. Experiment 2 was continued for two more epochs (a total of 3 epochs) of training with the limited resources available. The model tested using the dataset curated by \cite{rashid2024comprehensive}, with the resulting model having an accuracy of 94\% and a F1-score of 0.94. The model saw little to no changes in the negative sentiment classification performance. The detailed result of experiment 2 after 3 epochs can be found on Table \ref{table:ex2e3}.

As observed on Table \ref{table:ex1e3} and Table \ref{table:ex2e3}, the model trained on BanglishRev with rating information as sentiment label and tested on the dataset curated by \cite{rashid2024comprehensive} performs better than the baseline model trained and tested by the \cite{rashid2024comprehensive} dataset. This exemplifies the effectiveness of using rating information as a substitute for sentiment labels when training a sentiment analysis model. Interestingly, experiment 2 saw a slight decrease in scores after being trained for 2 additional epochs as can be observed in Table \ref{table:ex2e1} and Table \ref{table:ex2e3}. Further research may validate and expand upon these findings. 

\subsection{Limitations and Ethical Considerations} \label{limitations}
The data collected in the dataset is unbalanced when it comes to sentiment analysis or rating prediction tasks as more than 78\% of the reviewers gave 5 star ratings to the products. While this could have been balanced by removing 5 star rated reviews to have a better distribution over ratings, we refrained from doing so to keep the dataset open for other tasks such as spam review detection, market analysis or consumer behavioral pattern detection. It is difficult and computationally draining to train supervised model with the dataset due to the substantial size of the dataset. Working with a subset of the data may prove to be less computationally expensive.

Due to space limitations, the links to the images associated with the reviews were provided in the dataset instead of the actual images. We intended to upload a version of the images with reduced resolution with the dataset in the future. We do not control the image links provided in the dataset. Not all of those images might be available at a later point of time, it may be removed or replaced. 

The data for BanglishRev was collected from the publicly available reviews on the Daraz Bangladesh Ltd. All data in the dataset are extracted by the authors. The data has been anonymized with sensitive information like product IDs and user IDs being replaced with unique numbers that does not have any relationship to the original IDs. We strongly advice this dataset to be used for academic research and non-commercial purposes only. No human subjects or animals were involved in the collection of the data. Usage of the BanglishRev dataset may aid in the expansion of e-commerce in the context of Bengali population and result in the same negative social impacts that arise with this expansion. \cite{lubbe2002economic} discussed the potential positive and adverse effects that comes with the rapid expansion of e-commerce platforms.

Review data in the dataset was collected in a period between April, 2024 and May, 2024. BanglishRev is curated solely from Daraz and does not contain data from the other available platforms. While Daraz is the largest e-commerce platform in the region now, there are other platforms like FoodPanda/Pathao Food for food delivery, Rokomari for books etc. in specified domains thriving in their own sectors \citep{islam2022review}. On that account, the dataset may not provide a full picture of the broader e-commerce market of Bangladesh. We acknowledge this limitation and plan on making a dataset more informative and inclusive of the other platforms in the future.

\subsection{Future Research Avenues}
BanglishRev is a large dataset with wealth of information hidden within it. Aside from sentiment analysis of the reviews, this dataset can be utilized in various other sectors. While browsing through the dataset we noticed that there were instances where a product got a suspiciously high number of 5 star ratings in an extremely short amount of time while competing products were either without reviews or given suspiciously less ratings. Upon further scrutinizing the reviews posted within that extremely small amount of time, they appeared to be very similar in wording and length. There can potentially be cases of products being spammed with lower ratings to aid in the sell of a competing product, extensive research into those behavioral patterns is required to make any conclusive statements about it. \cite{lim2010detecting} have worked on detecting spam reviews from rating patterns in an English Review dataset, we believe this can be an potential future research avenue utilizing BanglishRev for the Bengali population as well.

Aside from that, we observed that Automotive \& Motorebike and the TV \& home appliances categories were the ones with the least amount of reviews, which may indicate that there are either lesser number of consumers in those categories or the consumers are less likely to leave a review. It may also be the case that for more expensive home appliances and automotive, the consumers are more likely to lean towards offline markets and stores. The BanglishRev dataset contains the sub category information alongside the 8 root categories, more such behavioral patterns can be analyzed in future works. The larger amount of data present in the dataset would make the patterns observed statistically significant. 

The large amount of positive reviews present in the dataset and the positive words that appeared in the word clouds for both Bangla and English may also indicate that most consumers in the e-commerce platform is satisfied with the products bought, or that consumers are more likely to leave a review if they are satisfied with the products. It may also be the case the negative ratings and reviews are deleted by the sellers. Future research can delve deeper into the observation by conducting qualitative analysis over subsections of the dataset.

The image data in the dataset can be utilized in observing correlations between customer satisfaction and the similarity of the image with the product thumbnail or image (Are customers more satisfied if the images are similar to the product images?) and if it differs for different cultures (Utilizing BanglishRev for Bangla and comparing against datasets from other regions to observe the patters across different cultures). It might be a possible research avenue to look for customer preferences or likeliness of purchasing a product related to the product in the review from the images associated with the reviews (by analyzing which products are observed together) without violating privacy of the customer (as the data is anonymized). Future research can also focus on annotating the dataset to utilize it for specific tasks.

\section{Conclusion} \label{Conclusion}
BanglishRev, an extensive and large scale review dataset focused on the Bengali population that surpasses the previously available datasets on this domain is presented in this work. It stands as the largest dataset available for e-commerce platforms targeting this population, encompassing 1.74 million written reviews from 3.2 million ratings across 128,000 products. The inclusion of extensive metadata such as reviewer ratings, posting dates, purchase dates, likes, dislikes, seller responses, and associated images added a rich layer of contextual information that enhances its utility for different comprehensive analysis. The dataset has been evaluated for sentiment analysis using ratings as indications of sentiment and achieved an impressive accuracy of 94\% and an F1 score of 0.94, validating the potential of BanglishRev in facilitating sentiment analysis in similar cases where the reviews are written in a mixture of the different written forms of expression that the Bengali population uses. 

Discussions have also been provided to shed light into the limitations and ethical considerations related to the creation, usage and distribution of the dataset. We encourage responsible usage of the BanglishRev dataset, and advice the usage to be limited to educational and non-commercial purposes, acknowledging the significance of the ethical implications. Various future research directions where this dataset can be utilized have been discussed and we recommend future research or application of BanglishRev to be conducted with care. The BanglishRev dataset provides valuable resources focused on the Bengali population to the Bengali NLP researchers, marketing researchers and HCI researchers alike, and we have expectations that future research utilizing this dataset will greatly contribute to the research communities.

\bibliographystyle{apalike}
\bibliography{ref}

\end{document}